\documentclass[10pt,twocolumn,letterpaper]{article}
\pdfoutput=1
 
\usepackage{wacv}
\usepackage{times}
\usepackage{epsfig}
\usepackage{graphicx}
\usepackage{amsmath}
\usepackage{amssymb}
\usepackage{xcolor}
\usepackage{float}
\usepackage{caption}
\usepackage{subcaption}
\usepackage{adjustbox}

\wacvfinalcopy 

\def\wacvPaperID{92} 

\ifwacvfinal\fi
\begin{document}

\title{Gender-From-Iris or Gender-From-Mascara? }

\author{Andrey Kuehlkamp \hspace{1cm} Benedict Becker \hspace{1cm} Kevin Bowyer \\
Department of Computer Science \& Engineering, University of Notre Dame\\
{\tt\small akuehlka@nd.edu, kwb@nd.edu}
}

\maketitle
\ifwacvfinal\thispagestyle{empty}\fi

\begin{abstract}
Predicting a person's gender based on the iris texture has been explored by several researchers.
This paper considers several dimensions of experimental work on this problem, including person-disjoint train and test, and the effect of cosmetics on eyelash occlusion and imperfect segmentation.
We also consider the use of multi-layer perceptron and convolutional neural networks as classifiers, comparing the use of data-driven and hand-crafted features.
Our results suggest that the gender-from-iris problem is more difficult than has so far been appreciated.
Estimating accuracy using a mean of N person-disjoint train and test partitions, and considering the effect of makeup - a combination of experimental conditions not present in any previous work - we find a much weaker ability to predict gender-from-iris texture than has been suggested in previous work.
\end{abstract}

\section{Introduction}

Classifying gender based on iris texture has been explored by several researchers, with a range of reported accuracies. 
Different features, classifiers and methods to evaluate accuracy have been used.
Although the results indicate that the iris texture contains information related to gender, no work to date has described the texture appearance that characterizes each gender. 

Neural Networks (NNs) are known as powerful classifiers, and for being able to autonomously learn features from the training data. 
Due to these properties, and to the current popularity of NN solutions in computer vision and biometrics, we explore the use of NNs for gender-from-iris.
Apart from the classifier, several ways of extracting image features can be used. 
The simplest is to use pixel intensities, but more sophisticated techniques may result in more powerful features.
We categorize feature extraction techniques into \textit{data-driven}, which are learned automatically by the NN classifiers, and \textit{hand-crafted}, that applies some specifically defined transformation over the raw data.

Most gender-from-iris work to date has overlooked one or more questions that may be important: 
\textit{What is the accuracy breakdown by gender?
Is gender-from-iris based on true iris texture differences, or based on incidental factors such as presence/absence of eye makeup?
How important is subject-disjoint training and testing in getting true performance estimates?
Do Convolutional Neural Networks (CNNs) offer any performance improvement over hand-crafted features and classifiers for gender-from-iris?
}

This paper describes results of experiments that  explore these questions.
We compare the use of Multi-Layer Perceptrons (MLPs) and CNNs for gender-from-iris.
We use different approaches to extract information from the iris texture; we analyze the accuracy achieved for each gender; we look into the bias that may be created by the use of cosmetics; and we look at the bias that results from not using a subject-disjoint training and testing.

\section{Related Works}

\begin{table*}[!htbp]
    \centering
    \resizebox{\textwidth}{!}{%
        {\renewcommand{\arraystretch}{2.5}%
        \begin{tabular}{ c|c|c|c|c|c|c|c|c } 
             \hline
             \textbf{Authors}    & \textbf{Classifier}    & \textbf{Accuracy(\%)}  & \textbf{Features}  & \textbf{Dataset Size} & \parbox{1.5cm}{\centering\textbf{Person-Disjoint}}  & \parbox{1.8cm}{\centering\textbf{Cross-Validation}} & \textbf{Cosmetics} & \parbox{2cm}{\centering\textbf{Breakdown by Gender}} \\ \hline
             
             \parbox{2cm}{\centering Thomas \etal \\(2007) \cite{Thomas2007}}    &  Decision Tree  &   75    & \parbox{2.5cm}{\centering Gabor filtering + \\Hand-Crafted}  & 57,137    & No    & 10f & No & No \\  \hline
             
             \parbox{2cm}{\centering Lagree \etal \\(2011) \cite{Lagree2011PredictingEA}} & SVM & 47.67--62.17 & \parbox{2.5cm}{\centering Gabor filtering + \\Hand-Crafted}    & 600   & Yes   & 2, 5 and 10f & No & No \\ \hline
             
             \parbox{2cm}{\centering Bansal \etal \\(2012) \cite{Bansal2012}} & SVM & 83.06  & \parbox{2.5cm}{\centering Hand-crafted + DWT} & 300   & No   &   10f & No & No \\  \hline
             
             \parbox{2cm}{\centering Tapia \etal \\(2014) \cite{Tapia2014}} & SVM & 96.67   & LBP   & 3,000 & No    & 80/20 & No & Yes \\  \hline
             
             \parbox{2.1cm}{\centering Fairhurst \etal \\(2015) \cite{Fairhurst2015}} & \parbox{2.1cm}{\centering Various (individual and combined)} & 49.61--89.74 & \parbox{2.6cm}{\centering Geometric+Texture Hand-Crafted}  & 1,600 & Yes   & 72/28  & No & No \\ \hline
             
             \parbox{2cm}{\centering Tapia \etal (2016) \cite{Tapia2016GenderCF}} & SVM   & 91; 85.33  & IrisCode  & 3,000; 3,000\footnotemark   & No; Yes    & 80/20  & No & Yes \\ \hline
             
             This paper &   MLP/CNN & $66\pm2.7$  & \parbox{2.5cm}{\centering Intensity, Gabor filtering, LBP}    & 3,000 & Yes   & 80/20, 10x    & Yes   & Yes \\
             \hline
        \end{tabular} %
        }
    }
    \caption{Overview of gender prediction from iris images.}
    \label{tab:relatedworks}
\end{table*} 

The extraction of ancillary information from biometric traits is known as \textit{soft biometrics}.  
As defined by Dantcheva \etal \cite{Dantcheva2016}, "[s]oft biometric traits are physical, behavioral, or material accessories, which are associated with an individual, and which can be useful for recognizing an individual."

Gender is one soft biometric attribute, and gender recognition has been explored using  biometric traits such as faces, fingerprints, gait and irises.
The earliest work on gender-from-iris \cite{Thomas2007} used a classifier based on decision trees, and reported an accuracy of about 75\%. 
They extracted hand-crafted geometric and texture features from log-Gabor-filtered images in a dataset of over  57,000 images. 
The training and testing sets were not person-disjoint, which typically results in a higher estimated accuracy than can be expected for new persons.

Later, \cite{Lagree2011PredictingEA} used a Support Vector Machine (SVM) classifier with features extracted using spot and line detectors and Law's texture measures. 
They used a dataset of 600 images and a cross-validation protocol with 2, 5 and 10 folds, with person-disjoint partitions. 
They considered both race-from-iris and gender-from-iris, and their classification accuracy on gender-from-iris ranged from $47\%$ to $62\%$. 
A similar approach was used by \cite{Bansal2012}, which used 2D Discrete Wavelet Transform (DWT) in combination with hand-crafted statistical features to extract texture information from the images.
Using an SVM to classify the irises, they reported accuracy up to $83\%$ on a small dataset of 300 images. 

In the work of \cite{Tapia2014}, using an SVM to classify Local Binary Pattern (LBP) features extracted from 3,000 iris images yielded an accuracy of $96.67\%$. 
This was for an $80/20$ $train/test$ split, on non-person-disjoint partitions. 
The same authors used a similar technique to perform gender classification based on the IrisCode used for identification in \cite{Tapia2016GenderCF}. 
In this work, they performed evaluation on two different datasets: one was person-disjoint, while the other was not, and the reported accuracy changed considerably. 
The person-disjoint dataset, called the Gender-from-Iris (GFI) dataset, is available to the research community.

In another study, \cite{Fairhurst2015} used an SVM in a combined consensus with other classifiers to achieve 81\% accuracy on a person-disjoint dataset. 
They used a combination of geometric and texture features, selected via statistical methods, and a $72/28$ training/testing split to prevent overfitting.

\begin{figure*}[!htbp]
    \captionsetup[subfigure]{justification=centering}
    \centering
    \begin{subfigure}[b]{0.45\textwidth}
        \centering
        \includegraphics[width=\textwidth]{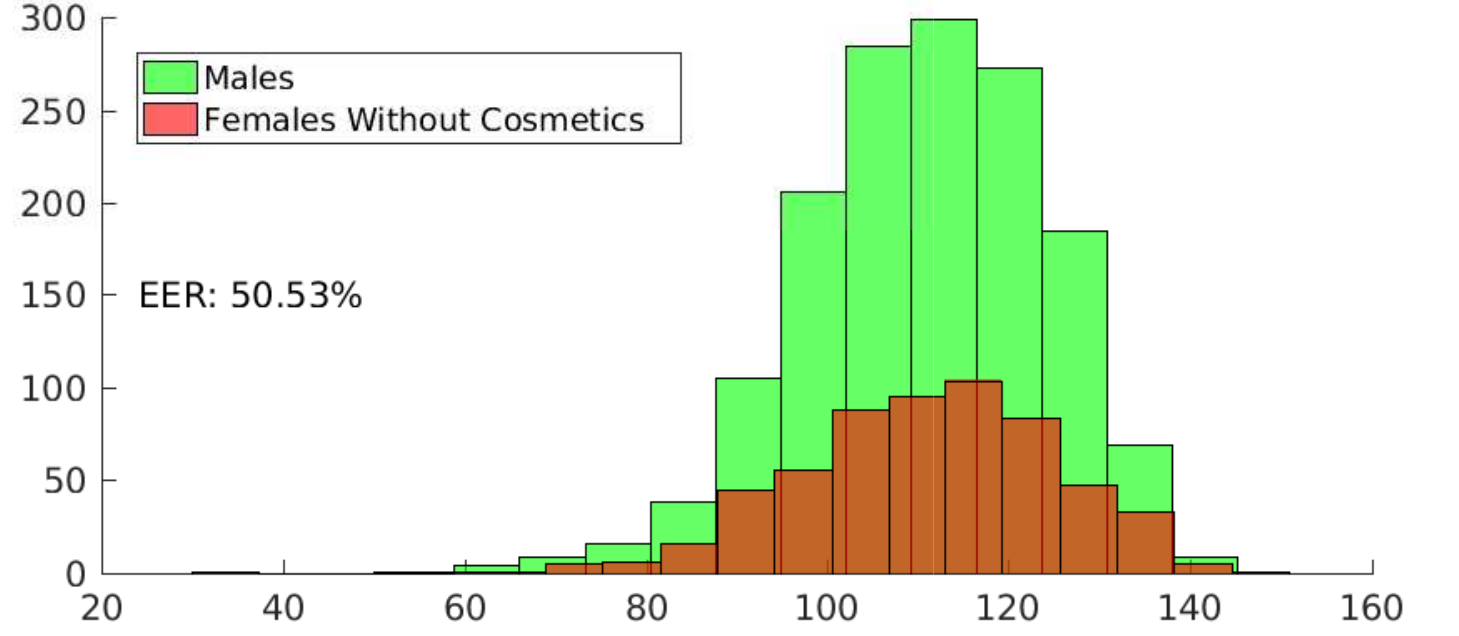}
        \caption{Males and \textit{Females with No Cosmetics}, whole eye images.}
        \label{fig:disteyefnc}
    \end{subfigure}
    \hfill
    \begin{subfigure}[b]{0.45\textwidth}
        \centering
        \includegraphics[width=\textwidth]{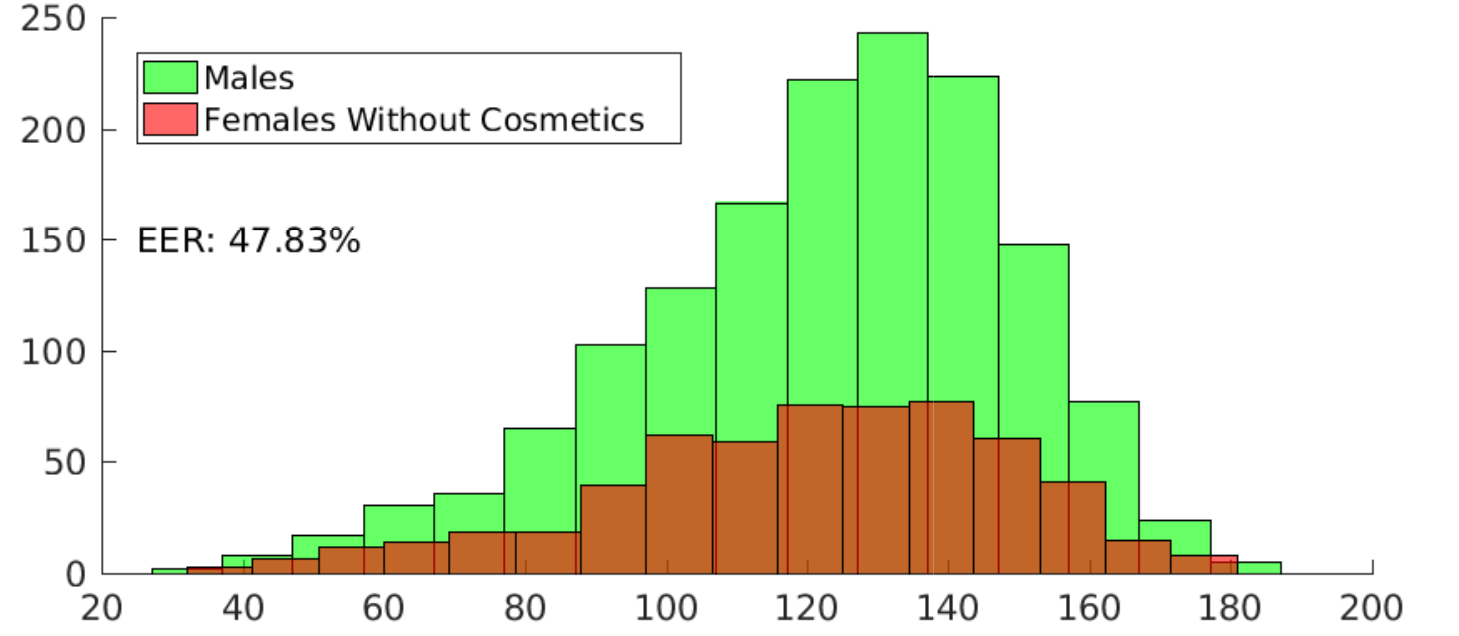}
        \caption{Males and \textit{Females with No Cosmetics}, segmented irises.}
        \label{fig:distirisfnc}
    \end{subfigure}
    \\
    \begin{subfigure}[b]{0.45\textwidth}
        \centering
        \includegraphics[width=\textwidth]{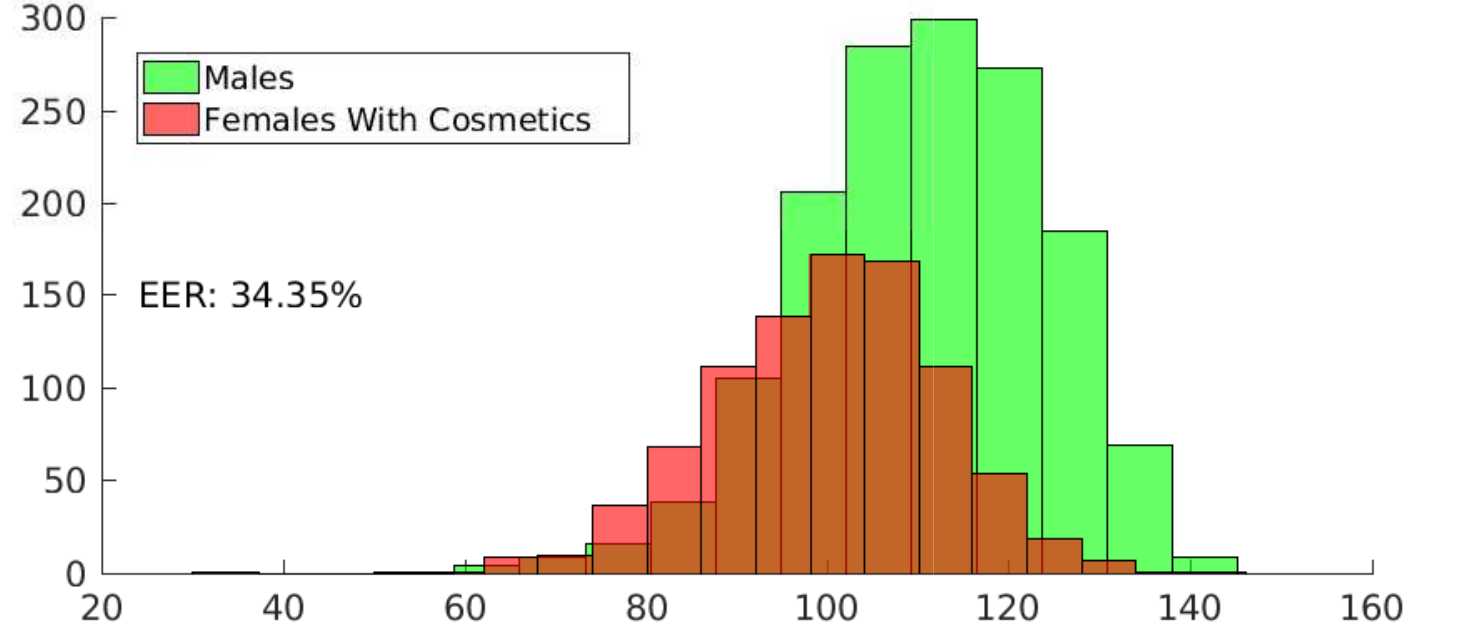}
        \caption{Males and \textit{Females With Cosmetics}, whole eye images.}
        \label{fig:disteyefwc}
    \end{subfigure}
    \hfill
    \begin{subfigure}[b]{0.45\textwidth}
        \centering
        \includegraphics[width=\textwidth]{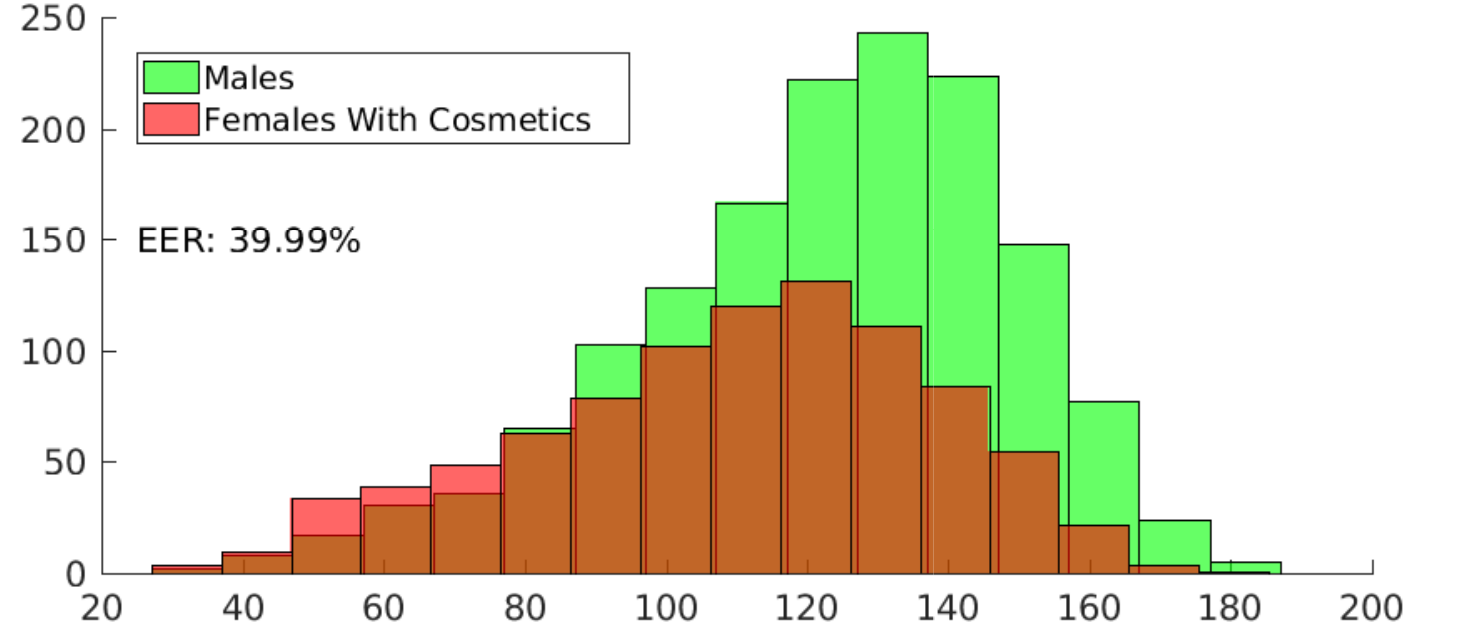}
        \caption{Males and \textit{Females Without Cosmetics}, segmented irises.}
        \label{fig:distirisfwc}
    \end{subfigure}
    \caption{Threshold on average image intensity can achieve $60\%$ correct classification of males and FWC.}
    \label{fig:eer}
\end{figure*}

\footnotetext{The first set of 3,000 was not person-disjoint, so the authors used another 3,000 images person-disjoint set.}

An overview of the techniques and results used so far is presented in Table  \ref{tab:relatedworks}.
None of these works has looked systematically at the effect of cosmetics 
on accuracy of predicting gender-from-iris.
Most of the works do not use a subject-disjoint training and testing, especially those reporting the highest accuracy.
And these works report accuracy from a single random split into train-test data, rather than a mean of N random splits.
Apart from \cite{Fairhurst2015}, no other research employed neural networks for this task.


\section{Methods}
\label{methods}

We use the "Gender from Iris" (GFI) dataset \footnote{https://sites.google.com/a/nd.edu/public-cvrl/data-sets} used in \cite{Tapia2016GenderCF}, which to our knowledge
is the only publicly available dataset for this problem.
It consists of 
$1,500$ left-eye and $1,500$ right-eye images, for $3,000$ total, representing 750 male and 750 female subjects. 
The $480\times640$, near-infrared images were obtained with an LG 4000 iris sensor.

Previous work generally reported accuracy based on a single random split of the data into train and test.
The problem with this is that a single partitioning of the data into train and test can easily give an "optimistic" estimate of true accuracy.
For this reason, in our experiments, a basic trial is a random 80/20 split into train and test data, 
and reported accuracy is averaged over ten trials. 
Each trial is person-disjoint training and testing. 
With this approach, we expect to obtain a more true estimate of accuracy.

The iris images were processed using IrisBee \cite{Liu2005} to segment and normalize the iris region. 
Normalized iris images were stored in different resolutions: $40\times240$, $20\times240$, $10\times240$, $5\times120$, $3\times60$ and $2\times30$ pixels. 
As a result of the segmentation, a mask is generated for each image, marking where the iris texture is occluded, usually by eyelids or eyelashes.
In the experiments that used raw pixel intensities as the features, the normalized iris images were used as feature inputs of the classifier. 
The sizes of the feature vectors were then $4800$, $2400$, $600$, $180$ and $60$, respectively. 

After performing training on a portion of the images, we use the test set to perform the evaluation, based on a simple criterion: given an unlabeled normalized iris, can we correctly predict the subject's gender? 
Two main feature extraction techniques were explored: data-driven features using raw pixel intensity, and hand-crafted features using Gabor filtering and LBP.
A more detailed description of these feature extraction approaches is given in section \ref{featext}.

Classification experiments were performed using MLP neural networks and CNNs. 
The details about the topology of the networks are described in section \ref{nntopo}.

\section{Person-Disjoint Train and Test}
\label{method_aspects}

We performed the same experiment on the person-disjoint GFI dataset, and on a previous version of that dataset that is not person-disjoint.
For the GFI dataset, there is one image per iris, and so the training and testing is necessarily person-disjoint.
For the second dataset, there are a varying number of images per iris, of a smaller number of different irises, and so the training and testing is not person-disjoint.
For both sets of results, accuracy is averaged over 10 trials, with each trial using a random 80/20 split for train/test data.

\begin{figure}[htb]
    \centering
    \includegraphics[width=1\linewidth]{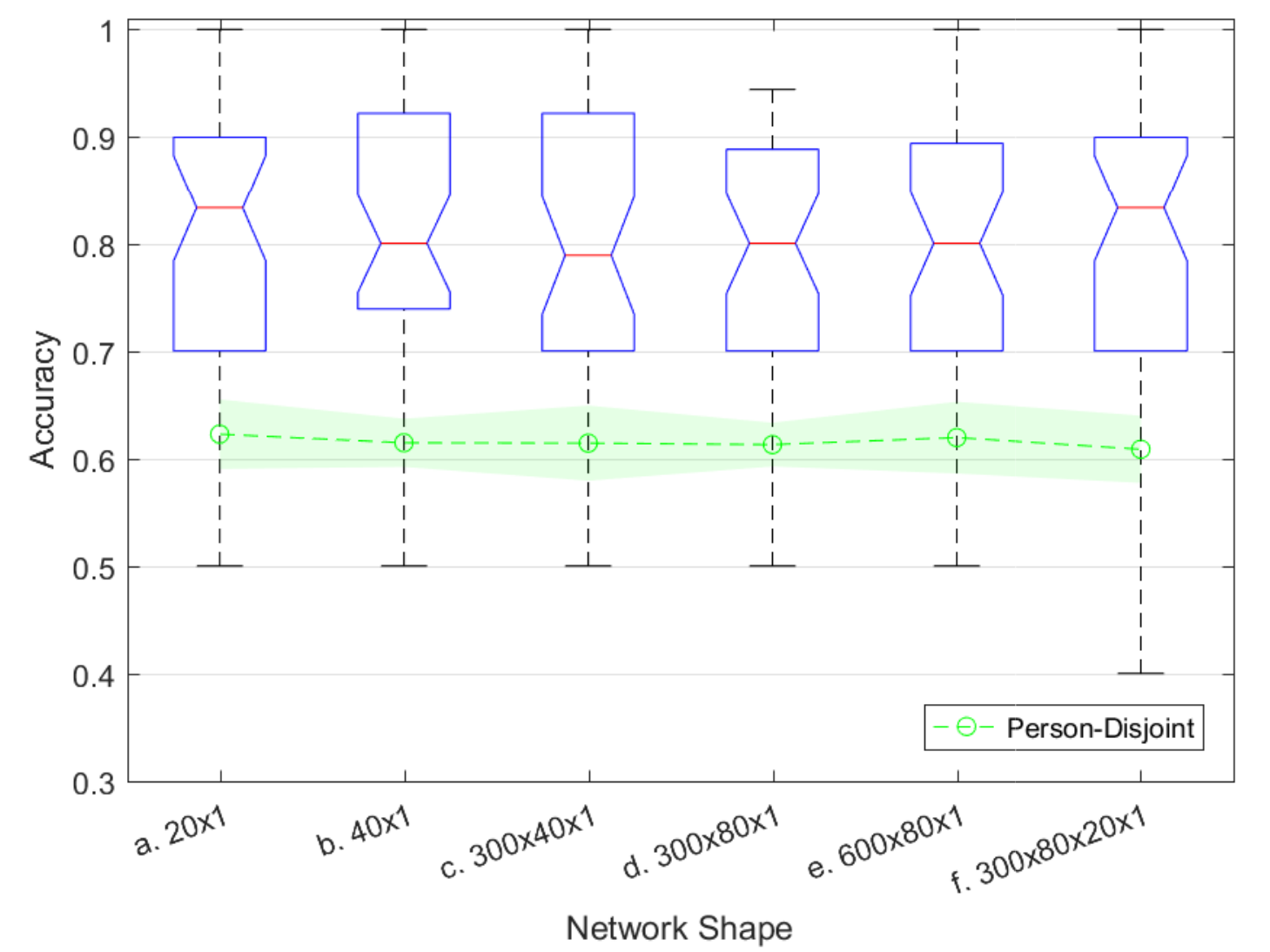}
    \caption{Box plots show accuracy distributions on a non-person-disjoint dataset, using an MLP classifier operating on raw pixel intensity. The green dotted line and shaded area represent the average accuracy and deviation on a person-disjoint dataset.}
    \label{fig:persondisjoint}
\end{figure}

The estimated accuracy using the subject-disjoint training and testing enforced by the GFI dataset is $61\% \pm 2.9$.
The estimated accuracy with the non-person-disjoint training and testing allowed by the other dataset with multiple images per iris is $77\% \pm 2.6$.
This is an average over ten trials; Figure \ref{fig:persondisjoint} shows that a single non-person-disjoint trial could easily result in an estimated accuracy of 100\%.
The higher estimated accuracy for the non-person-disjoint train/test apparently results from the classifier learning subject-specific features, rather than generic gender-related texture features.

This experiment makes the point that it is impossible to meaningfully compare non-person-disjoint results with person-disjoint results.
Higher (but optimistic) accuracies are reported for works using a non-subject-disjoint methodology and lower (but more realistic) accuracies reported using a subject-disjoint methodology.
Also, in general, accuracies are reported for a single split of the data.
A more useful accuracy estimate is computed over N trials using random person-disjoint splits of the data.

\section{Male/Female or Mascara/No Mascara?}
\label{sec:mascara}

\begin{figure}[!ht]
    \captionsetup[subfigure]{justification=centering}
    \centering
    \begin{subfigure}[b]{0.49\linewidth}
        \centering
        \includegraphics[width=\textwidth]{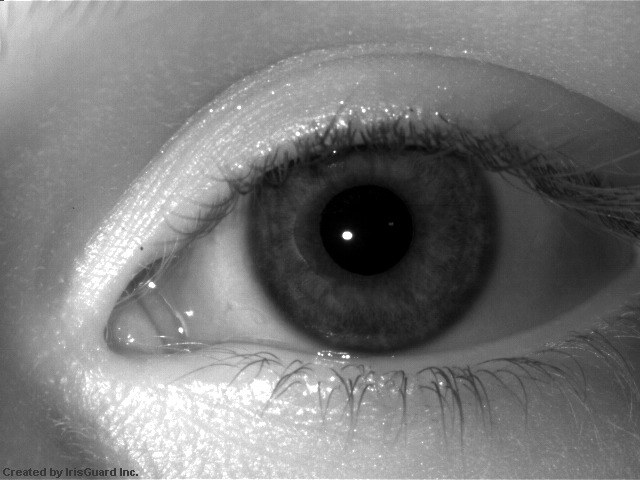}
        \caption{Eye without mascara}
        \label{fig:eyenomasc}
    \end{subfigure}
    \hfill
    \begin{subfigure}[b]{0.49\linewidth}
        \centering
        \includegraphics[width=\textwidth]{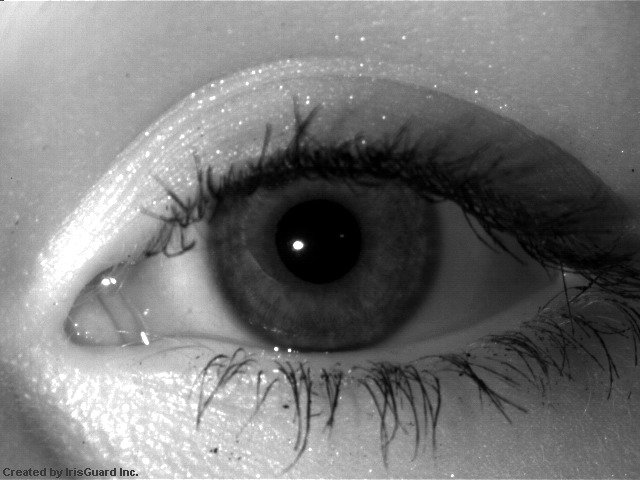}
        \caption{Eye with mascara}
        \label{fig:eyemasc}
    \end{subfigure}
    \\
    \begin{subfigure}[b]{0.49\linewidth}
        \centering
        \includegraphics[width=\textwidth]{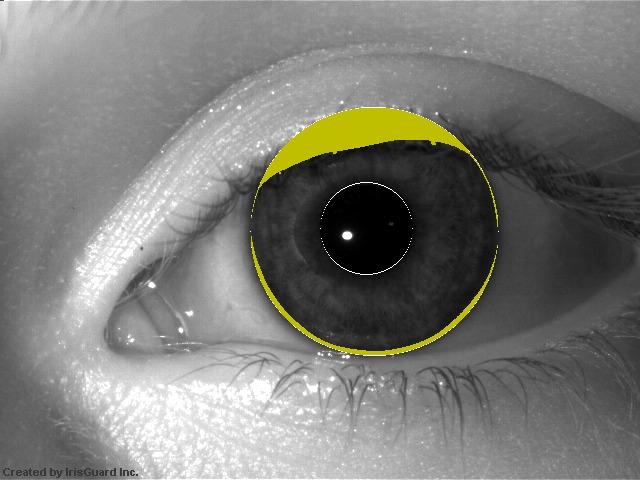}
        \caption{Segmented eye without mascara}
        \label{fig:eyesegwo}
    \end{subfigure}
    \hfill
    \begin{subfigure}[b]{0.49\linewidth}
        \centering
        \includegraphics[width=\textwidth]{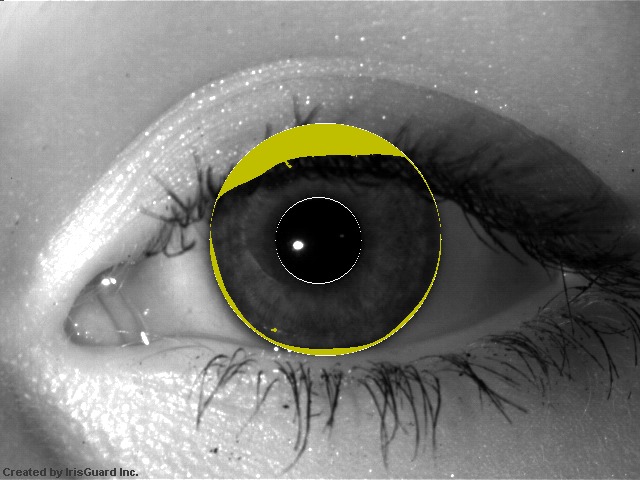}
        \caption{Segmented eye with mascara}
        \label{fig:eyesegwith}
    \end{subfigure}
    \\
    \begin{subfigure}[b]{0.49\linewidth}
        \centering
        \includegraphics[width=\textwidth]{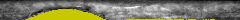}
        \caption{Normalized eye without mascara}
        \label{fig:eyenormwo}
    \end{subfigure}
    \hfill
    \begin{subfigure}[b]{0.49\linewidth}
        \centering
        \includegraphics[width=\textwidth]{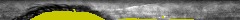}
        \caption{Normalized eye with mascara}
        \label{fig:eyenormwith}
    \end{subfigure}
    \\
    \begin{subfigure}[b]{0.49\linewidth}
        \centering
        \includegraphics[width=\textwidth]{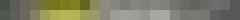}
        \caption{Resized normalized eye without mascara}
        \label{fig:reseyenormwo}
    \end{subfigure}
    \hfill
    \begin{subfigure}[b]{0.49\linewidth}
        \centering
        \includegraphics[width=\textwidth]{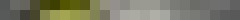}
        \caption{Resized normalized eye with mascara}
        \label{fig:reseyenormwith}
    \end{subfigure}
    \caption{Segmentation and normalization process in two images of the same eye, with and without mascara.}
    \label{fig:eyemascara}
\end{figure}

Mascara causes the eyelashes to appear thicker and darker in the iris image.
Figure \ref{fig:eyemascara} shows a female eye with and without mascara.
The use of eye makeup has been shown to affect iris recognition accuracy \cite{Doyle2013}.
The basic mechanism is that if eyelash segmentation is not perfect, the segmented iris region may include some eyelash occlusion.
To the degree that eyelash occlusion is present in the iris region, the use of mascara will generally increase the magnitude of the artifact in the texture computation.
The same effect can also happen with other types of makeup like eyeliner, although this one is applied to the eyelid instead of the eyelashes.

To investigate how mascara might affect gender-from-iris results, we reviewed the GFI dataset and annotated which images show evidence of mascara or eyeliner.
Just over $60\%$ for the female iris images show visible evidence of cosmetics, compared to 0\% of the male iris images.
The annotation allowed us to perform experiments using three categories of images: Male, Female With Cosmetics (FWC) and Female with No Cosmetics (FNC). 

One simple observation is that average image intensity for FWC is darker than for FNC or for Males (Fig. \ref{fig:eer}).
This is true whether one considers the image as a whole, or only the segmented iris region.
For Males and FNC, the distributions of average image intensity are almost identical; see Fig. \ref{fig:disteyefnc} and \ref{fig:distirisfnc}. 
For Males and FWC, there is a noticeable separation between the distributions; see Fig. \ref{fig:disteyefwc} and \ref{fig:distirisfwc}.
Based on this separation, we could apply a simple threshold and achieve better than $60\%$ accuracy distinguishing Males from FWC (EER of about $37\%$).
However, a similar threshold for Males and FNC results in only about $50\%$ accuracy.
This experiment shows how the presence of mascara can potentially make the gender-from-iris problem appear to be easier to solve than it is in reality.

\begin{figure*}[!htbp]
    \captionsetup[subfigure]{justification=centering}
    \centering
    \begin{subfigure}[b]{0.3\textwidth}
        \centering
        \includegraphics[width=\textwidth]{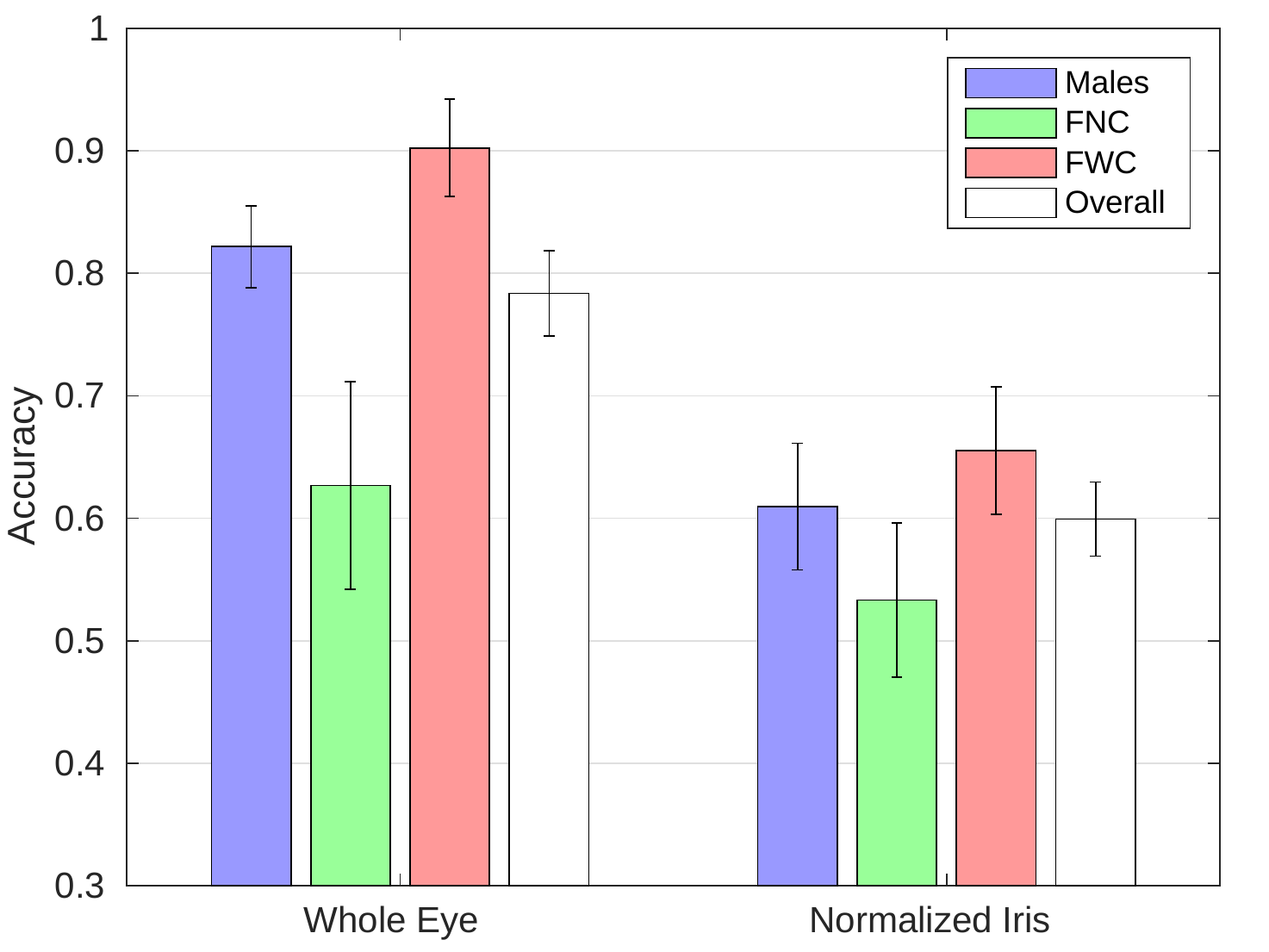}
        \caption{Training on \\All subjects}
        \label{fig:irisall}
    \end{subfigure}
    \hfill
    \begin{subfigure}[b]{0.3\textwidth}
        \centering
        \includegraphics[width=\textwidth]{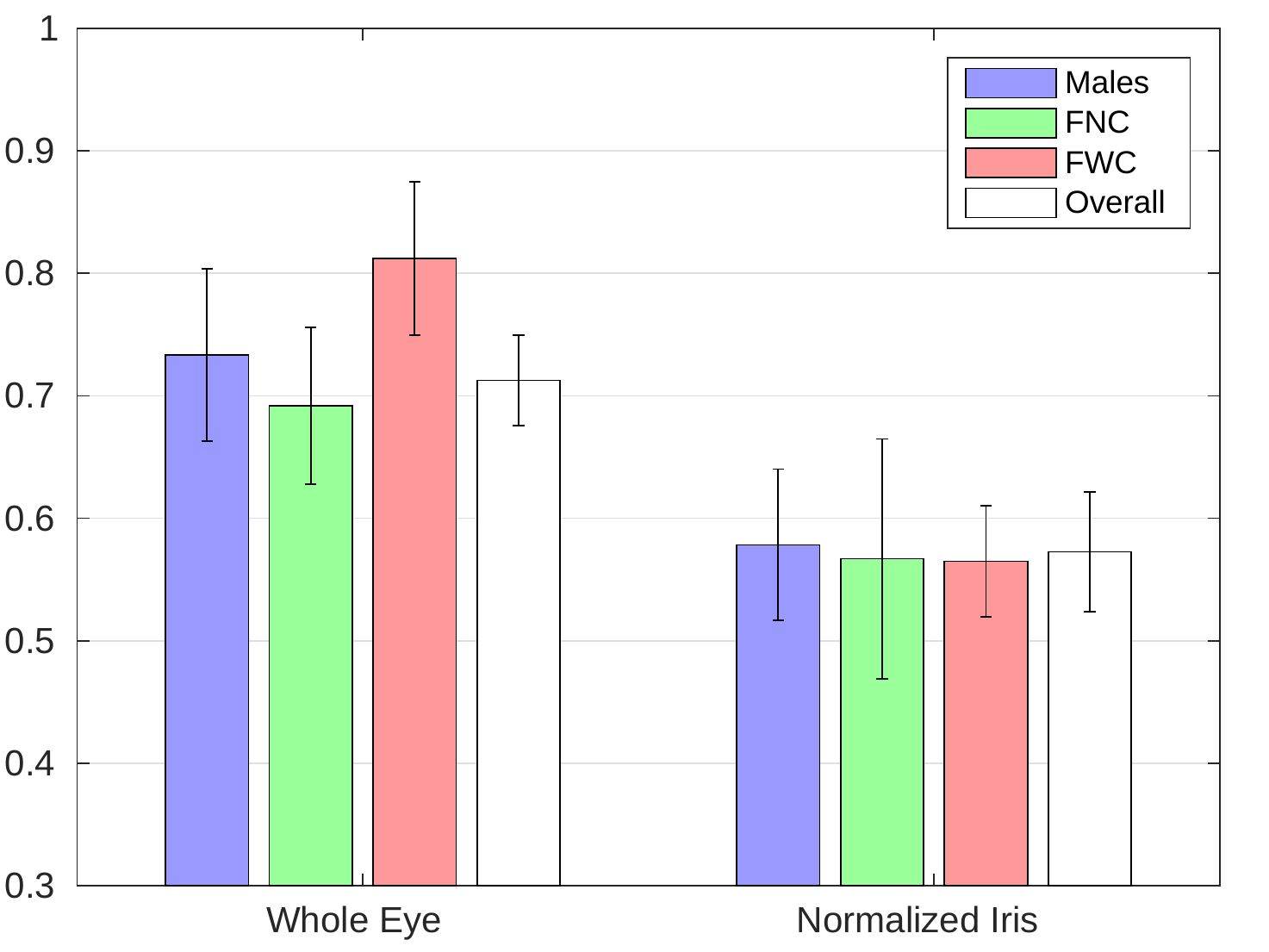}
        \caption{Training on \\Males+Females with No Cosmetics}
        \label{fig:irisfnc}
    \end{subfigure}
    \hfill
    \begin{subfigure}[b]{0.3\textwidth}
        \centering
        \includegraphics[width=\textwidth]{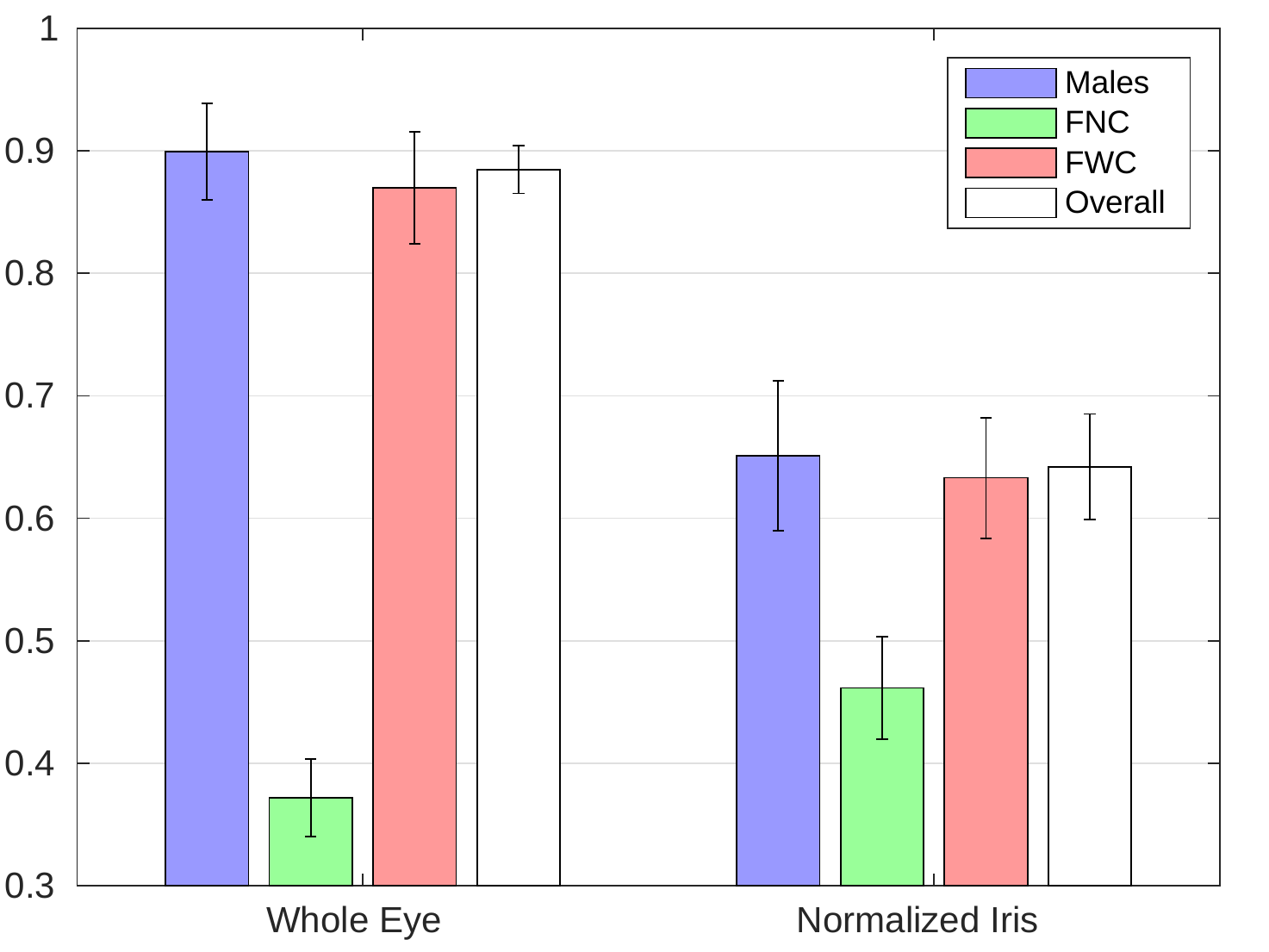}
        \caption{Training on \\Males+Females With Cosmetics}
        \label{fig:irisfwc}
    \end{subfigure}
    \caption{Gender classification accuracy for different training groups, using pixel intensity.}
    \label{fig:irisdist}
\end{figure*}

We also trained MLP networks to classify gender-from-iris. 
We considered both using the whole iris image, and using only the normalized iris region.
We also considered training with and without images containing mascara. 
The results are summarized in Figure \ref{fig:irisdist}.
When training with the full dataset (Males, FNC and FWC), the accuracy achieved with the whole image is greater than the accuracy achieved with the iris region alone.
Also, the accuracy achieved is highest for the FWC subgroup, and lowest for the FNC subgroup.
The trained MLP is apparently able to use the presence of mascara to correctly classify a higher fraction of the females in the FWC subgroup, at the expense of lower classification for the FNC subgroup.

Next we trained two additional networks, one using Males + FNC, and another using Males + FWC.
The Male images were randomly sampled to equal the number of female images, to avoid biasing the training toward a majority class.
Comparing the results for normalized iris trained on all subjects (Fig. \ref{fig:irisall} right) with those trained on Males+FNC (Fig. \ref{fig:irisfnc} right), while FNC performance improved, we can perceive a small decrease in the overall accuracy.
At the same time, training on Males+FWC (Fig. \ref{fig:irisfwc} right) causes the overall performance to increase to $64\%$. 

This effect is amplified when working with whole eye images.
In Fig. \ref{fig:irisfnc} (left side) the FNC accuracy improvement is almost equal to the male accuracy drop, and it results in an overall accuracy contraction with regard to Fig. \ref{fig:irisall} (left).
On the other hand, in Fig. \ref{fig:irisfwc} (left) the overall accuracy rises to $88\%$. 

The experiment makes it clear that mascara is an important confounding factor for gender-from-iris.
If mascara is present in the dataset, then it is hard to know the degree to which the classifier learns gender from iris texture versus gender from mascara.
Future research on gender-from-iris should use datasets that include annotations for the presence of mascara, and new mascara-free datasets are needed.

\section{Occlusion masks}
\label{sec:irisocclusions}

Eyelids and eyelashes frequently occlude portions of the iris,
Ideally, the segmentation step would result in these occlusions becoming part of the "mask" for the image.
Results in the previous section indicate that eyelash occlusion is generally not perfectly segmented. 
It appears that mascara causes the "noise" resulting from un-masked eyelash occlusion to become a feature that can be correlated with gender.
If this is the case, mascara may also cause more eyelash occlusion to be identified and segmented (Fig. \ref{fig:eyemascara}).
In this case, the size and shape of the masked region would be a feature correlated with gender.

\begin{figure}[htb]
    \centering
    \includegraphics[width=1\linewidth]{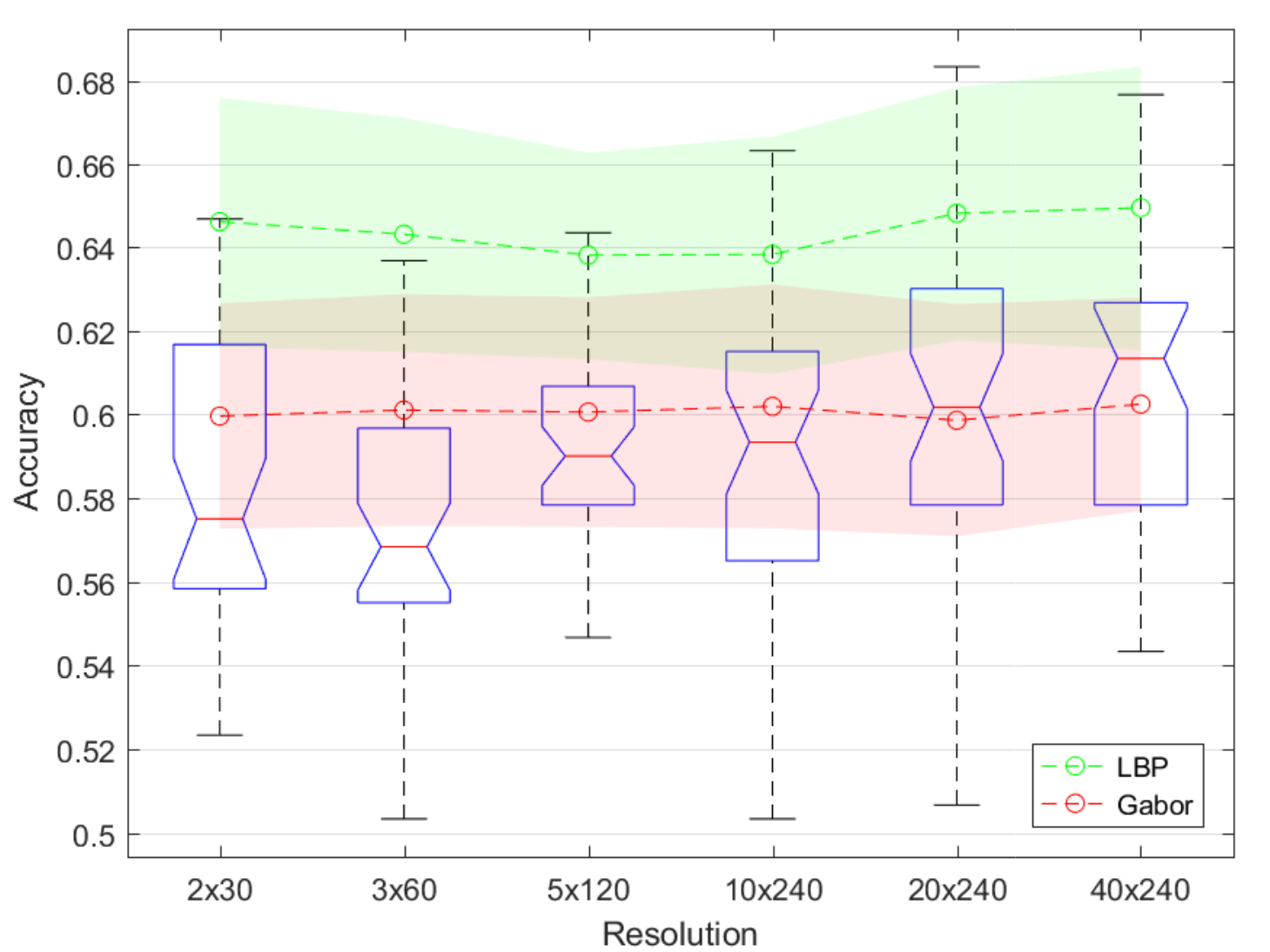}
    \caption{Accuracy using only the binary occlusion mask of the normalized iris. The dotted green line and shaded area denote the average accuracy and deviation for classifying normalized irises using LBP, and the dotted red line and shaded area are average accuracy and deviation using Gabor filtering. The box plots are accuracy distribution using simply the occlusion masks for the same normalized irises.
}
    \label{fig:maskcomp}
\end{figure}

In order to determine if the occlusion mask contains gender-related information,
we performed an experiment where the only information given to the MLP classifier is the (binary) occlusion mask. 
Figure \ref{fig:maskcomp} shows the result of this experiment.
Despite the fact that the MLP has no access to any iris texture information, the accuracy achieved is  similar to that achieved on the iris images. 

The results of this experiment suggest that there are two paths by which mascara can make it easier to identify female iris images.
To the degree that eyelash occlusion is not well segmented, the eyelash occlusion that contaminates the iris texture will be darker with mascara than it is without.
To the degree that mascara makes it easier to segment more of the eyelash occlusion, the masked area of the iris will be larger.
By whichever path, when high gender-from-iris accuracy is found using a dataset in which many women wear mascara, it is difficult to know if the accuracy is truly due to gender-related patterns in the iris texture, or simply due to the presence of mascara.

\section{Features: hand-crafted and data-driven}
\label{featext}

Approaches explored to extract discriminative features from the normalized iris images include  hand-crafted features (e.g., Gabor filtering, LBP) and data-driven features, in which the raw pixel intensity is fed into neural networks that may "learn" features.
All the experiments followed the same methodology, described in Section \ref{methods}. 

\subsection{Data-Driven Features}

Neural networks are an example of a classifier that can learn features from raw data.
Data-driven features are "learned" from a dataset through a training procedure, and are dependent on the characteristics of the data and the classification goal. 
Here we present results of this approach, obtained through MLP and CNN classifiers. Details on the implementation of these networks are in Section \ref{nntopo}.

Pixel intensity is the simplest feature.
The pixel values of the masked, normalized image are fed directly to the neural network.
Despite no explicit texture information being given to the network, the average accuracy of this approach was approximately $60\%$. 
This accuracy is similar to what could be achieved using a simple intensity thresholding on the images, as seen in Section \ref{sec:mascara}.
This suggests that in this instance the neural network may be learning to predict gender based on a measure of average pixel intensity, or some other feature that is no more powerful.

\begin{figure}[htb]
    \centering
    \includegraphics[width=1\linewidth]{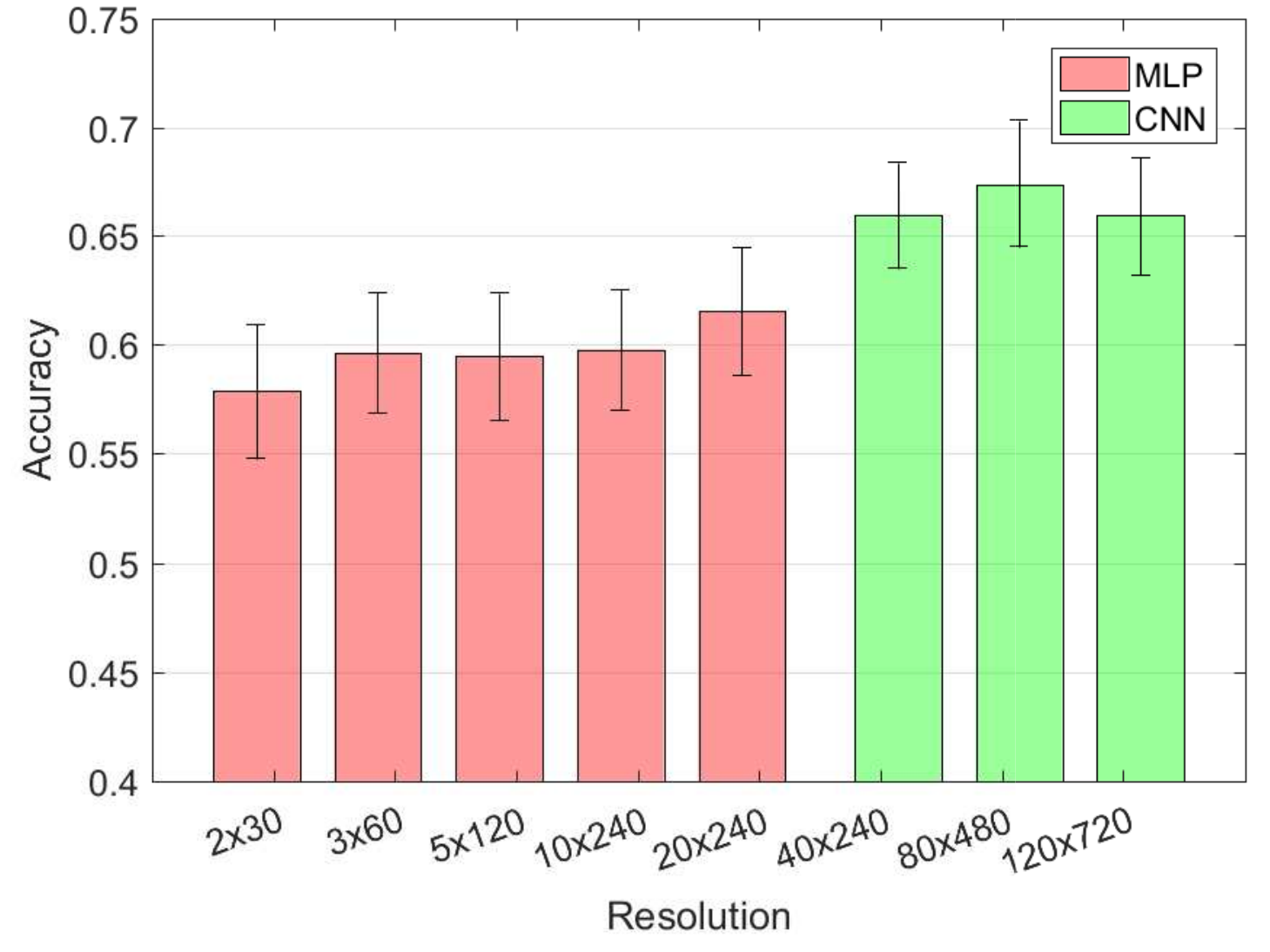}
    \caption{Accuracy based on pixel intensity for MLP and CNN classifiers. Average and standard deviation across 10 randomized train/test repetitions.}
    \label{fig:iris_raw_resolution}
\end{figure}

Figure \ref{fig:iris_raw_resolution} shows a plot of the average accuracy obtained by this technique, across different image resolutions.
It is worth observing that low resolutions like 2x30 and 3x60 the images could contain very little texture information because of the averaging of pixels. 

\subsection{Hand-Crafted Features}

Gabor filtering and LBP are popular examples of hand-crafted feature extraction techniques.
Gabor filtering is done as part the standard approach to creating the "iris code" \cite{Daugman2004,Daugman2016}.
In our experiments here, 1-Dimensional Gabor filtering was performed for each row of the normalized iris.
We chose to explore a range of wavelengths similar to those used for iris recognition.
LBP has been used in previous work on gender-from-iris \cite{Tapia2014}.

For Gabor-filtered iris images, the average accuracy was $57\% \pm 3$ across all wavelengths, and there was no significant difference between different wavelengths considered. 
The fact that Gabor filtering resulted in  worse classification than pixel intensity may seem surprising, but there is a possible explanation.
Gabor filtering highlights the local occurrence of certain frequencies in the image by maximizing its response to these frequencies, while minimizing the response to other frequencies. 
If these low-response frequencies are related to features like occlusions or mascara, it makes sense that its attenuation has a negative effect on accuracy. 
As shown in section \ref{sec:mascara}, the presence of eye cosmetics or even occlusion masks may artificially enhance the gender classification accuracy, and their removal makes the problem harder.
So these results are consistent with the idea that a significant part of the information that is used for gender classification may not come from the iris texture.

It is also important to mention that this work was limited to testing a certain range of parameters, based on those used for iris recognition. 
Since the main objective of iris recognition is to maximize the distinction between individual subjects and attenuate all non-person-specific features (such as gender, race, eye color, etc.), these parameters may not be the most appropriate for gender classification. 

Local Binary Patterns (LBP) is a well-known method for texture analysis \cite{Ojala2002,Guo2010}.
We took some of the same LBP variations and parameters in \cite{Tapia2014}, and used MLP neural networks to perform gender prediction.

\begin{figure}[htb]
    \centering
    \includegraphics[width=1\linewidth]{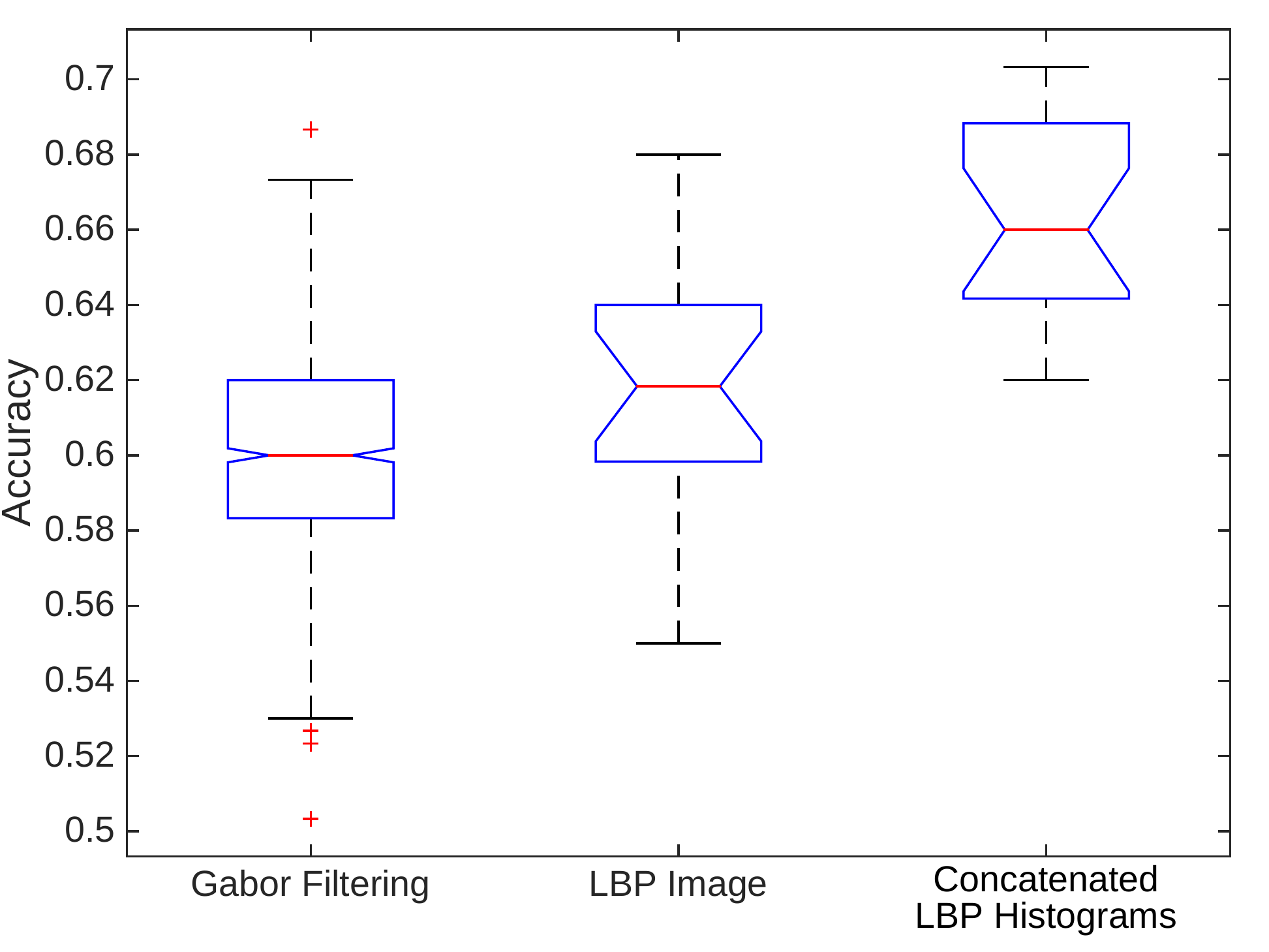}
    \caption{Classification accuracy distributions for hand-crafted features in a MLP classifier.}
    \label{fig:featext}
\end{figure}

In general, the best performances were achieved by uniform patterns and its variations (ULBP, CULBP-Mag and CULBP-Sign). 
ULBP histograms with and without patch overlapping had the highest accuracy with an average of $66\%$.
Figure \ref{fig:featext} shows an overall comparison between the three different feature extraction techniques. 
Gabor filtering had the worst results, with an average accuracy a little above $58\%$. 
In this graph, LBP extraction is divided into two different categories because of the significant performance difference between them. 
LBP images yields better accuracy than Gabor filtering or pixel intensity, but still well below Concatenated LBP Histograms.

\section{Neural Network Topologies}
\label{nntopo}

It is difficult to characterize the specific geometric or texture features that can be used to distinguish male from female irises. Thus, we decided to use an approach based on neural networks, so that they could learn the features that are best fit for this classification.

The first portion of these experiments consists of an exploratory attempt to classify gender, training arbitrary-sized MLP Neural Networks using backpropagation. 
As a rule of thumb for the structuring of the networks, all of them had a first hidden layer of $1.5 \times P$, where $P$ is the number of input features. 
The following layers of neurons in the network were defined as shown in Table \ref{tab:annshape}. 
For example for a $10 \times 240$ image, the first network was configured as $3600 \times 20 \times 1$, the second $3600 \times 40 \times 1$, and so on.

In the cases where resolutions higher than 20x240 were used, the size of the MLP had to be reduced due to memory limitations. In these cases, we limited the size of Layer 1 to 5,000 neurons.

\begin{table}[H]
    \centering
    \resizebox{\columnwidth}{!}{
    \begin{tabular}{c|c|c|c|c}
        \hline
        Layer 1 & Layer 2 & Layer 3 & Layer 4 & Output \\ \hline
        $P \times 1.5$ & 20 & - & - & 1 \\
        $P \times 1.5$ & 40 & - & - & 1 \\
        $P \times 1.5$ & 300 & 40 & - & 1 \\
        $P \times 1.5$ & 300 & 80 & - & 1 \\
        $P \times 1.5$ & 600 & 80 & - & 1 \\
        $P \times 1.5$ & 300 & 80 & 20 & 1 \\ 
        \hline
    \end{tabular}
    }
    \caption{MLP neural network topologies used. $P$ is the number of input pixels.}
    \label{tab:annshape}
\end{table}

The activation function used for each layer of the network was a hyperbolic tangent, with the exception of the output layer, which consisted of a sigmoid activation function, in order to produce an output within the range of 0 and 1 corresponding to the gender.

\begin{figure}[!htbp]
    \centering
    \includegraphics[width=\linewidth]{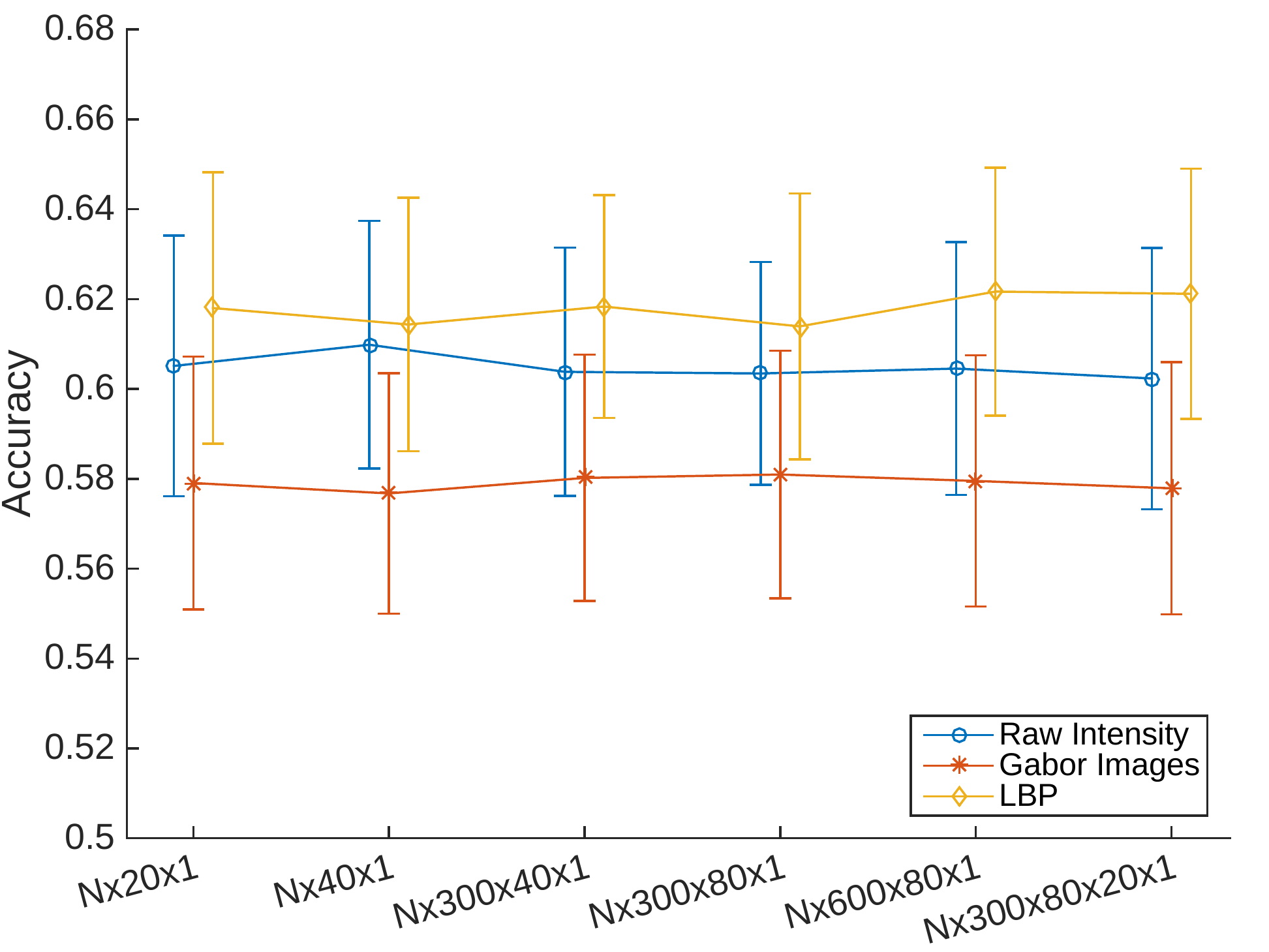}
    \caption{Accuracy of MLP classifier on Gender prediction, using different features according to network topology. Average and standard deviation across 10 randomized repetitions.}
    \label{fig:acccomparison}
\end{figure}

\begin{figure*}[!htbp]
    \centering
    \includegraphics[width=\textwidth]{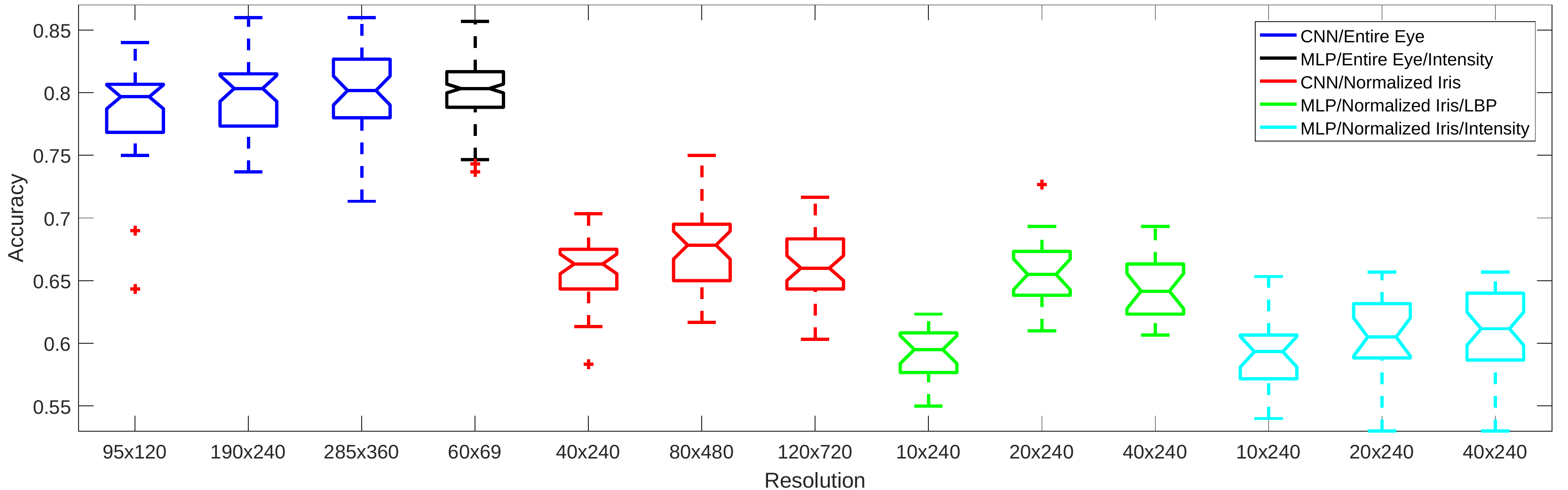}
    \caption{Comparison of CNN with MLP results. Blue and black boxes represent accuracy using the entire eye image, respectively for CNN and MLP. Red, green and cyan boxes denote the accuracy using the normalized iris image.}
    \label{fig:fullimagecnn}
\end{figure*}

Network topology, within the range of options explored here, seems to have very little effect on the classification accuracy. 
Figure \ref{fig:acccomparison} shows how little variation occurs across different topologies with different types of image features.
These results also emphasize that LBP features perform better than raw intensity or Gabor features.

\subsection{Convolutional Neural Networks}
\label{subcnn}

We also experimented with a CNN architecture. These architectures have seen great progress in prominent image recognition benchmarks \cite{VGG}, \cite{googlenet}, and their success in N-way image classification makes them  promising for binary image classification as well. For the purposes of this paper, a CNN was used to classify gender based on two inputs: the full image and the segmented iris image with black occlusion masks (Figure \ref{fig:fullimagecnn}, blue and red plots). 

While the networks described in \cite{VGG} and \cite{googlenet} are extremely large, trial by experimentation and difficulty of task (1000-way multi-scale classification vs 2-way single-scale classification) led to the conclusion that a smaller architecture would suffice in this environment. The network used consists of 3 sets of CNN layers, followed by 2 fully-connected (FC) layers and a softmax output. Each CNN set consisted of a Convolutional layer with a $(4,4)$ kernel and a $(1,1)$ stride, followed by a Max Pooling layer with a $(2,2)$ kernel and a $(2,2)$ stride. The number of features in each CNN layer were 16, 32 and 64 respectively, and the number of neurons in the FC layers were 1024 and 1536. Each neuron in the CNN and FC layers used the activation function $max(0,X)$, commonly known as a Rectified Linear Unit, or ReLU activation. 

Like before, GFI data was randomly split into $80/20$ person-disjoint subsets for training/testing, and the network trained on 2500 batches of 32 images before testing in all cases. 
The training was carried out separately for left and right eyes on three different resolutions: $95\times120$, $190\times240$ and $285\times360$ for the entire eyes, and $40\times240$, $80\times480$ and $120\times720$ for normalized irises. 
Twenty randomized trials for each resolution and eye were performed.

Surprisingly, the results were virtually the same for all eyes and resolutions and almost identical to the accuracies obtained from using MLP networks. 
This may be because the data embedded in the image is low-level and separable by the MLP network, so the CNN layers simply transfer the underlying data to the final FC layers instead of extracting more information through its convolutions. 
This phenomenon would result in similar accuracies across network topologies and input resolutions, like those produced in this paper's experiments.

If we look at the resolutions used in the experiments with CNN and MLP on the entire eyes (Fig. \ref{fig:fullimagecnn}, blue and black boxes), the lower resolution used with MLP shows there is no accuracy gain using larger images, or using a more complex classifier.
This means that classification is relying on image blobs that are large enough to be detected in a $60 \times 69$ image, once again suggesting that the fine details of iris texture do not contribute to gender classification as much as it was initially thought.
When we look comparatively to normalized iris images (Fig. \ref{fig:fullimagecnn}, red, green and cyan boxes), a significant portion of the gender-related information is lost.
Again, CNNs do not seem to have a substantially higher accuracy.



\section{Conclusions}

We showed how the use of non-person-disjoint training and test can result in estimated gender-from-iris accuracy that is biased high.
We also showed the importance of averaging over multiple trials.
Using a single random train-test split of the data, the estimated accuracies ranged from $40\%$ to $100\%$.

We showed that the presence of eye makeup results in 
higher estimated gender-from-iris accuracy.
We also showed that classification based on the occlusion masks, disregarding completely the iris texture, results in an accuracy of approximately $60\%$.
And we showed that simple averaging of the iris image intensity and thresholding can result in approximately $60\%$ gender-from-iris accuracy.

Our experiments showed hand-crafted features like LBP to yield better prediction accuracy ($66\%$) than data-driven features ($60\%$) when using MLP networks.
On the other hand, CNNs (using data-driven features) had performance comparable to MLPs+LBP.
In a similar experiment using the entire eye images, CNNs and MLPs had equivalent performance (around $80\%$) using learned features.

Previous research may have misjudged the complexity of gender-from-iris, especially because of the subtle but important factors explored here. 
For future work, we suggest the creation of a subject-disjoint, mascara-free dataset. 
Currently, it is not clear what level of gender-from-iris accuracy is possible based solely on the iris texture.

\section{Acknowledgements}

The authors thank Dr. Adam Czajka for his invaluable insight and contribution.

This research was partially supported by the Brazilian Ministry of Education -- CAPES through process BEX 12976/13-0. 

{\small
\bibliographystyle{ieee}
\bibliography{egbib}
}

\end{document}